# 中文分词十年又回顾: 2007-2017[*]

# Chinese Word Segmentation: Another Decade Review (2007-2017)

赵海  Hai Zhao
蔡登  Deng Cai
上海交通大学  Shanghai Jiao Tong University
黄昌宁  Changning Huang
清华大学  Tsinghua University
揭春雨  Chunyu Kit
香港城市大学  City University of Hong Kong

**[内容简介]**

本文回顾中文分词在2007-2017十年间的技术进展，尤其是自深度学习渗透到自然语言处理以来的主要工作。我们的基本结论是，中文分词的监督机器学习方法在从非神经网络方法到神经网络方法的迁移中尚未展示出明显的技术优势。中文分词的机器学习模型的构建，依然需要平衡考虑已知词和未登录词的识别问题。尽管迄今为止深度学习应用于中文分词尚未能全面超越传统的机器学习方法，我们审慎推测，由于人工智能联结主义基础下的神经网络模型有潜力契合自然语言的内在结构分解方式，从而有效建模，或能在不远将来展示新的技术进步成果。

**Abstract**

This paper reviews the development of Chinese word segmentation (CWS) in the most recent decade, 2007-2017. Special attention was paid to the deep learning technologies that has already permeated into most areas of natural language processing (NLP). The basic view we have arrived at is that compared to traditional supervised learning methods, neural network based methods have not shown any superior performance. The most critical challenge still lies on balancing of recognition of in-vocabulary (IV) and out-of-vocabulary (OOV) words. However, as neural models have potentials to capture the essential linguistic structure of natural language, we are optimistic about significant progresses may arrive in the near future.

**[关键词]**
中文分词 (Chinese word segmentation)  神经网络 (neural networks)

# 1 背景

中文分词是中文信息处理的一个基础任务和研究方向。十年前，黄和赵(2007)接受《中文信息学报》委托，针对自20世纪末以来的中文分词的机器学习方法做了十年回顾，发表了《中文分词十年回顾》一文。该文的基本结论是中文分词的统计机器学习方法优于传统的规则方法，尤其在未登录词（out-of-vocabulary words, OOV)即训练集之中未出现的词的识别上，具有无可比拟的优势。这一基本结论随后得到全面证实。现在10年过去，截至2017年3月，Google Scholar显示该文被引用166次，而中国知网记录其引用为483次。

今天看来，使用机器学习方法在具有切分标记的分词语料上学习或训练出高效能中文分词器是自然而然的想法，然而十年前的情形大不相同。首先，直到20世纪的最后十年，中文信息处理界才意识到分

词可以作为真实标注语料上的机器学习任务进行操作。其次，充足的语料准备也并非一蹴而就。两个早期的语料来自宾州大学中文树库(Chinese Penn Treebank, CTB) 和北京大学计算语言所标注的人民日报语料。在各类切分语料齐备的基础上，SIGHAN才得以组织第一次国际性的中文分词评测SIGHAN Bakeoff-2003。

语料准备之外，还有两个历史性的因素，迟滞了中文分词这一中文信息处理基础子任务走向彻底的机器学习。其一，长期以来，中文分词的经典方法，即最大匹配算法，在合适的词典搭配下，通常能够取得一定程度上颇可接受的性能。以F值度量，最大匹配分词一般情况下约能获得80%甚至更高的成绩。这类简单有效的规则方法的存在，极大降低了研发先进机器学习技术的迫切性。其二，机器学习方法的计算代价巨大，在必要硬件条件尚未普及或者成本仍过于高昂的情况下，机器学习方法的优势得不到体现。2005年，典型的分词学习工具条件随机场（conditional random field, CRF）在百万词语料库上的训练，需要12-18小时的单线程CPU时间，占用内存2-3G，远超当时个人计算机的一般硬件配置水准。

因此，我们在2017年的今天回顾10年前的学术状态，必须历史性地考虑当时当地的具体情形，才能理解当时以及后来那些理所当然的技术进步有其内在而特殊的合理性和必然性。最近10年机器学习领域最为显著的技术井喷，显然是深度学习方法的崛起和全面覆盖。因而，我们下面将技术总结分为两大部分，即中文分词的传统机器学习模型和最近的深度学习（神经网络）模型。本文对技术论文的最新引用截止至ACL-2017会议的部分已录用论文。但由于篇幅所限，我们仅关注严格意义上的监督机器学习模式下的相关工作，而对于非监督学习、半监督学习、领域迁移学习以及其它分词方法和应用等，则尚付阙如，有待日后或各路高贤的努力。我们冒昧以此相对狭窄的视角，回顾我们力所能及范围内的一些当时及当今相对前沿的研究工作，旨在抛砖引玉，以供借鉴。

## 2 传统的机器学习模型

分词作为字符串上的切分过程，是一种相对简单的结构化机器学习任务。根据所处理的结构分解单元，大体可以将用于分词的传统机器学习模型分为两大类，即基于字标注的和基于词（相关特征）的学习。

基于字标注学习的方法始于Xue (2003)。该工作使用一个字在词中的四种相对位置标签(tag)，即B、M、E和S等字位（如表1所示），来表达该字所携带的切分标注信息，从而首次将分词任务形式化为字位的串标注学习任务。串标注学习是自然语言处理中最基础的结构化学习任务，在串标注的概率图模型中，两个串的各个节点单元需要严格一一对应，非常方便于使用各种成熟的机器学习工具来建模和实现。Xue (2003)的首次实现其实尚未充分使用串标注结构学习，而是直接应用了字位分类模型。Ng & Low (2004)和Low et al. (2005)才是第一次将严格的串标注学习应用于分词，用的是最大熵（Maximum Entropy，ME）Markov模型。而Peng et al. (2004)和Tseng et al. (2005)则自然地将标准的串标注学习工具条件随机场引入分词学习。随后，CRF多个变种构成了深度学习时代之前的标准分词模型。

表1: Xue (2003)的字位标注示例

| 自 | 然 | 科 | 学 | / | 的 | / | 研 | 究 | / | 不 | 断 | / | 深 | 入 |
|---|---|---|---|---|---|---|---|---|---|---|---|---|---|---|
| B | M | M | E | | S | | B | E | | B | E | | B | E |
| B：开始，M：中间，E：结束，S：单字词 | | | | | | | | | | | | | | |

中文分词任务是切分出特定上下文环境下正确的词，因此，所谓基于词的分词学习建模需要解决一个"先有鸡还是先有蛋"的问题。基于词的随机过程建模引致一个CRF变种，即semi-CRF(半条件随机场)模型的直接应用。基于字位标注的分词学习通常用到的是线性链条件随机场(linear chain CRF)，它是基于Markov过程建模，处理过程中的每步只对输入序列的一个文本单元进行标注。而semi-CRF则基于semi-Markov过程建模，它在每步给序列中的连续单元标注成相同标签。这一特性和分词处理步骤高度契合，使其可以直接用于分词处理。

Andrew (2006)发表semi-CRF的第一个分词实现。然而，即使以当时的标准，号称直接建模的semi-CRF

模型的分词性能却不甚理想。通常来说，直接建模会获得更好的机器学习效果，然而在semi-CRF直接应用于分词时，却一直很难兑现。之后，Sun et al. (2009)和Sun et al. (2012)将包含隐变量的semi-CRF学习模型用于分词，才将其分词性能提升到前沿水平：前者是首个隐变量semi-CRF模型的工作，声称能够同时利用基于字序列和基于词序列的特征信息，并经验证明引入隐含变量能通过有效捕捉长距信息来提升长词的召回率；后者额外引用了新的高维标签转移Markov特征，同时针对性地提出了基于特征频数的自适应在线梯度下降算法，以提升训练效率。值得注意的是，线性链CRF模型的训练时间比对应的最大熵Markov模型会慢数倍，因为最大熵模型训练时间正比于需要学习的标签数量，而CRF训练时间则正比于标签数量的平方，但semi-CRF的训练比标准的CRF还要缓慢，因此极大地限制了该类模型的实际应用。

传统的字标注模型方法在进一步发展之后，也引入部分标志性的已知词信息(即词表词，in-vocabulary words, IV)。Zhang et al. (2006)提出了一种基于子词（subword）的标注学习，基本思路是从训练集中抽取高频已知词构造子词词典。然而，该方法单独使用效果不佳，需要结合其他模型，其性能才能和已有方法进行有意义的比较。Zhao & Kit (2007a)大幅度改进了这个策略，通过在训练集上迭代最大匹配分词的方法，找到最优的子词（子串）词典，使用单一的子串标注学习即可获得最佳性能。

基于子串的直接标注模型事实上过强地应用了已知词信息，因为所有子串都属于已知词，并且在模型一开始就不能再切分。这一缺陷后来得到修正，主要的工作包括Zhao & Kit (2007b; 2008b; 2008a; 2011)。在这些工作中，对已有工作的改进主要有两点：其一，所有可能子串按照某个特定的统计度量方式根据训练集上的n-gram计数来进行打分；其二，基本模型还是字位标注学习，前面获得的子串信息以附加特征形式出现。这一工作获得了传统标注模型下的最佳性能，包括囊括2008年SIGHAN Bakeoff-4的全部五项分词封闭测试的第一名(Zhao & Kit, 2008b)。当子串的抽取和统计度量得分计算扩展到训练集之外，Zhao & Kit (2011)实际上提出了一种扩展性很强的半监督分词方法，实验也验证了其有效性。

和以上所有基于串标注，无论是线性链CRF标注还是semi-CRF标注的方法都不相同，Zhang & Clark (2007)引入了一种基于整句切分结构学习的分词方法。虽然他们声称这是一种基于词的方法，但是他们的方法不同于以往的最显著点，是字和词的n-gram特征，都以同等地位在整句的切分结构分解中进行特征提取。在细节上，他们采用了扩展的感知机算法进行训练，在解码阶段则使用近似的定宽搜索（beam search）。尽管其模型具备理论上更广泛的特征表达能力，但事实上该工作未能给出更佳的分词性能（参见表6的结果对比）。

由于分词是自然语言处理的一个起始任务，因此串标注学习下的可选特征类型相当有限。实际上，能选用的只是滑动窗口下的n-gram特征，n-gram单元为字或者词。理论上，以单个n-gram特征为单位进行任意的特征模板选择，在工程计算量上是可行的。实际的系统中，对于字特征多采取5字的滑动窗口，而Zhao et al. (2006a)及其后续工作则仅用3字窗口；对于词，则多采取3词的滑动窗口。然而，字位标注并非直接的分词学习，从后者（切分点）到前者（字位标注系统）有着多种方案，而一旦字位标注发生改变，相应的优化n-gram特征集显然会发生改变。这一现象的发现及其完善的经验研究，发表在Zhao et al. (2006b)和Zhao et al. (2010a)中。表2和表3分别列出了之前的标注集和Zhao et al. (2010a)考察过的完整标注序列，后者证明在6-tag标注集配合使用3字窗口的6个n-gram特征(分别是$C_{-1}$, $C_0$, $C_1$, $C_{-1}C_0$, $C_0C_1$, $C_{-1}C_1$，其中$C_0$代表当前字)，即可获得字位标注学习的最佳性能（默认使用CRF模型）。

表2: 早期的各类字位标签集

| 4-标签集 Low et al. (2005) Xue (2003) | | 3-标签集 Zhang et al. (2006) | | 2-标签集 Peng et al. (2004) Tseng et al. (2005) | |
|---|---|---|---|---|---|
| 词中字位 | 标签 | 词中字位 | 标签 | 词中字位 | 标签 |
| 开始 | B (LL) | 开始 | B | 开始 | Start |
| 中间 | M (MM) | 中间或结束 | I | 非开始 | NoStart |
| 结束 | E (RR) | | | | |
| 单字 | S (LR) | 单字 | O | | |

表3: 6-标签以下所有可能的字标签集及示例

| 标签集 | 标签 | 词中的字标注示例 |
|---|---|---|
| 2-tag | B, E | B, BE, BEE, … |
| 3-tag/a | B, E, S | S, BE, BEE, … |
| 3-tag/b | B, M, E | B, BE, BME, BMME, … |
| 4-tag | B, M, E, S | S, BE, BME, BMME, … |
| 5-tag | B, $B_2$, M, E, S | S, BE, $BB_2$E, $BB_2$ME, $BB_2$MME,… |
| 6-tag | B, $B_2$, $B_3$, M, E, S | S, BE, $BB_2$E, $BB_2B_3$E, $BB_2B_3$ME,… |

## 3 深度学习：神经网络分词模型

自从词嵌入（word embedding）表示达到了数值计算的实用化阶段之后，深度学习开始席卷自然语言处理领域。原则上，嵌入向量承载了一部分字或词的句法和语义信息，应该能带来进一步的性能提升。

如前所述，中文分词任务中可用的特征仅限于滑动窗口内的n-gram特征。由此，虽然典型的深度学习模型皆以降低特征工程代价的优势而著称，但是对于分词任务的特征工程压力的缓解却相当有限。因而，期望神经分词模型带来进一步性能改进的方向在于：一，有效集成字或者词的嵌入式表示，充分利用其中蕴含的有效句法和语义信息；二，将神经网络的学习能力有效地和已有的传统结构化建模方法结合，如在经典的字位标注模型中用等价的相应网络结构进行置换。

Collobert et al. (2011)提出使用神经网络解决自然语言处理问题，尤其是序列标注类问题的一般框架，这一框架抽取滑动窗口内的特征，在每一个窗口内解决标签分类问题。在此基础上，Zheng et al. (2013)提出神经网络中文分词方法，首次验证了深度学习方法应用到中文分词任务上的可行性。他们的工作直接借用了Collobert模型的结构，将字向量作为系统输入，其技术贡献包括：一，使用了大规模文本上预训练的字向量表示来改进监督学习(开放测试意义)；二，使用类似感知机的训练方式取代传统的最大似然方法，以加速神经网络训练。就结构化建模来说，该工作等同于Low et al. (2005)的字位标记的串学习模型，区别仅在于是用一个简单的神经网络模型替代了后者的最大熵模型，其模型框图见图1中的左图。由于结构化建模的缺陷，该模型的精度仅和早期Xue (2003)的结果相当，而远逊于传统字标注学习模型的佼佼者。

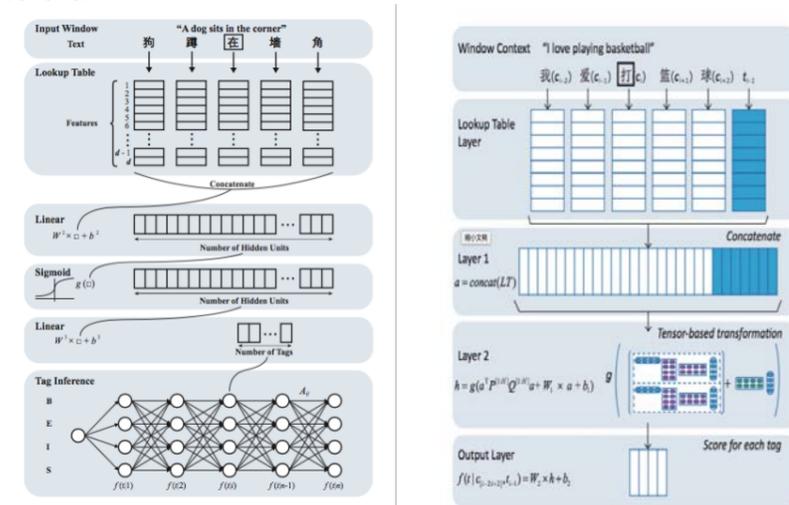

图 1: Zheng et al. (2013) (左)和Pei et al. (2014) (右)的模型框架。

2014年，Pei et al. (2014)对Zheng et al. (2013)的模型做了重要改进，引入了标签向量来更精细地刻画标签之间的转移关系，其改进程度类似于Low et al. (2005)首次引入Markov特征到Ng & Low (2004)的最大熵模型之中。Pei et al.提出了一种新型神经网络即最大间隔张量神经网络（Max-Margin Tensor Neural Network, MMTNN）并将其用于分词任务(见图1右)，使用标签向量和张量变化来捕捉标签与标签之间、

标签与上下文之间的关系。另外，为了降低计算复杂度和防止过拟合（所有神经网络模型的通病），该文还专门提出了一种新型张量分解方式。

随后，为了更完整精细地对分词上下文建模，Chen et al.（2015a）提出了一种带有自适应门结构的递归神经网络（Gated recursive neural network, GRNN）抽取n-gram特征，其中的两种定制的门结构（重置门、更新门）被用来控制n-gram信息的融合和抽取。与前述两项研究中简单拼接字级信息不同，该模型用到了更深的网络结构，为免于传统优化方法所受到的梯度扩散的制约，该工作使用了有监督逐层训练的方法。

同年，Chen et al. (2015b)针对滑动窗口的局部性，提出用长短期记忆神经网络（Long Short-Term Memory Neural Networks, LSTM）来捕捉长距离依赖，部分克服了过往的序列标注方法只能从固定大小的滑动窗口抽取特征的不足。Xu & Sun（2016）将GRNN和LSTM联合起来使用。该工作可以看作是结合了Chen et al. (2015a)和Chen et al. (2015b)两者的模型。该模型中，先用双向LSTM提取上下文敏感的局部信息，然后在滑动窗口内将这些局部信息用带门结构的递归神经网络融合起来，最后用作标签分类的依据。LSTM是神经网络模型家族中和线性链CRF同等角色的结构化建模工具，随着它被引入分词学习，神经网络模型在分词性能上开始可以和传统机器学习模型相抗衡。我们将结构化建模的传统-神经模型的对照情况列在表4中。

表4: 传统模型vs.神经模型：结构化建模演进（每行展示结构化建模对应的两种模型）

| 结构分解 | 传统模型 | 神经模型 |
| --- | --- | --- |
| 分类模型 | Xue (2003) | |
| Markov模型 | Ng and Low (2004); Low et al. (2005) | Zheng et al. (2013) <br> Pei et al. (2014) |
| 标准串学习建模 | CRF: Peng et al. (2004) <br> semi-CRF: Andrew (2006); Sun et al. (2009) | LSTM: Chen et al. (2015b) <br> Liu et al. (2016) |
| 全局模型 | Zhang and Clark (2007) | Cai and Zhao (2016) <br> Cai et al. (2017) |

与传统方法中基于字的序列标注方案几乎一统江湖的局面不同，神经网络有相对更灵活的结构化建模能力，因而有别于序列标注的其他方法也相继涌现出来。Ma & Hinrichs (2015)提出了一种基于字的切分动作匹配算法，该算法在保持相当程度的分词性能的同时，有着不亚于传统方法的速度优势。具体来说，该文提出了一种新型的向量匹配算法，可以视为传统序列标注方法的一种扩展，在训练和测试阶段都只有线性的时间复杂度(见图2左)。该工作有两个亮点值得注意：一，首次严肃考虑了神经模型分词的计算效率问题；二，遵循了严格的SIGHAN Bakeoff封闭测试的要求，只使用了简单的特征集合，而完全不依赖训练集之外的语言资源。

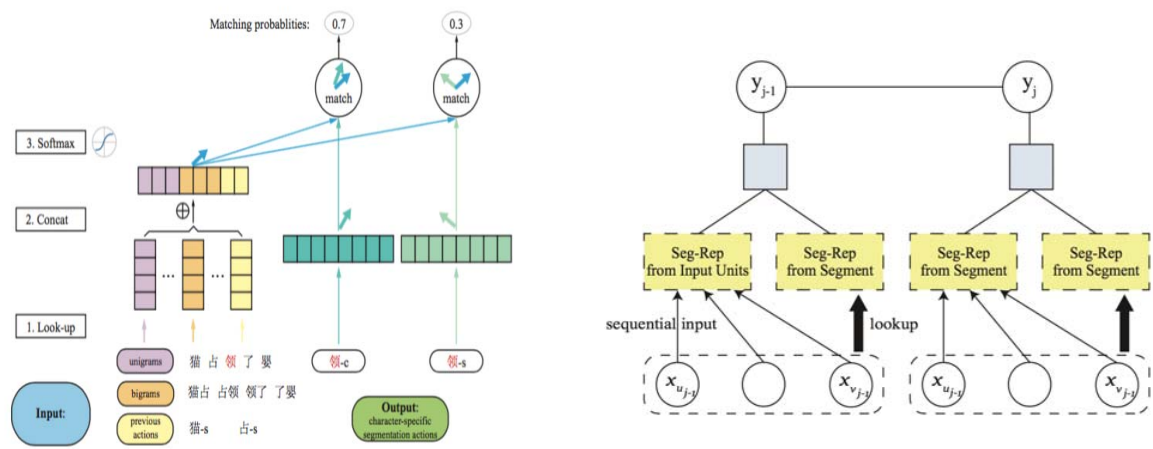

图2: Ma & Hinrichs (2015) (左)和Liu et al. (2016) (右)的模型框图

Zhang et al. (2016)提出了一种基于转移的模型用于分词，并将传统的特征模版和神经网络自动提取的特征结合起来，在神经网络自动提取的特征和传统的离散特征的融合方法做了尝试。结果表明，通过组合这两种特征，分词精度可以得到进一步提升。

Liu et al. (2016)首次将零阶半马尔可夫随机场应用到神经分词模型中，并分析了不同字向量和词向量对分词效果的影响。此文基于semi-CRF建模分词学习(见图2右)，用直接的切分块嵌入表示和间接的输入单元融合表示来刻画切分块，同时还考察了多种融合方式和多种切分块嵌入表示。遗憾的是，该系统严重依赖传统方法的输出结果来提升性能。他们的具体做法是用传统方法的分词结果（在外部语料上）作为词向量训练语料，因此，该文所报告的最终结果应属于开放测试范畴。而作为纯粹的神经模型版本下的semi-CRF模型，在封闭测试意义下，该系统的效果和传统semi-CRF（如Andrew (2006)）同样效果不佳（具体见表6的结果对照）。

Cai & Zhao(2016) 彻底放弃滑动窗口，提出对分词句子直接建模的方法，以捕捉分词的全部历史信息，提出了一个类似于Zhang & Clark (2007)的神经分词模型，同时充分吸收了前面一些工作的有益经验，如门网络结构等(图3)。由于覆盖了前所未有的特征范围，该模型在封闭测试意义上取得了和传统模型接近的分词性能。概括来说，该方法使用了一个带自适应门结构的组合神经网络，词向量表示通过其字向量生成，并用LSTM网络的打分模型对词向量序列打分。这种方法直接对分词结构进行了建模，能利用字、词、句三个层次的信息，是首个能完整捕捉切分和输入历史的方法。与之前的无论传统和还是深度学习的方法相比，该模型将分词动作依赖的特征窗口扩张到最大程度（见表5）。该文所提的分词系统框架可以分为三个组件：一个依据字序列的词向量生成网络组合门网络 (gated combination neural network，GCNN，见图4左)；一个能对不同切分从最终结果（也就是词序列）上进行打分的估值网络；和一种寻找拥有最大分数的切分的搜索算法。第一个模块近似于模拟中文造词法过程，这对于未登陆词识别有着重要意义；第二个模块从全句的角度对分词的结果从流畅度和合理性上进行打分，能最大限度地利用分词上下文；第三个模块则使在指数级的切分空间中寻找最可能的切分最优解。

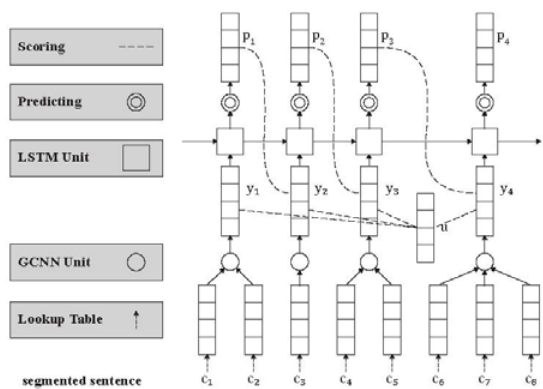

图 3: Cai & Zhao (2016)的模型框图

表5：不同模型的特征模板范围，其中i/j皆指当前打分的字或词

| 模型类别 | | 字特征 | 词特征 | 标签 |
|---|---|---|---|---|
| 基于字的 | Zheng et al. (2013), … | $c_{i-2}, c_{i-1}, c_i, c_{i+1}, c_{i+2}$ | - | $t_{i-1}t_i$ |
| | Chen et al. (2015b) | $c_0, c_1, …c_i, c_{i+1}, c_{i+2}$ | - | $t_{i-1}t_i$ |
| 基于词的 | Zhang and Clark (2007), … | c in $w_{j-1}, w_j, w_{j+1}$ | $w_{j-1}, w_j, w_{j+1}$ | - |
| | Cai and Zhao (2016) | $c_0, c_1, …, c_i$ | $w_0, w_1, …, w_j$ | - |

　　表6列出了近10年来主要的分词系统在SIGHAN Bakeoff-2005语料上的分词性能比较。神经分词系统短短数年间取得了长足进步，但整体上仍然不敌传统模型。此外，尽管神经网络方法在知识依赖和特征工程方面有着巨大优势，也取得了一定的进展，但模型的计算复杂度也大幅提高，因为成功的神经分词器往往建立于更加精巧、更复杂的网络结构之上。事实上，经历五年，深度学习方法在最终模型的性能上，无论是分词精度还是计算效率上，和传统方法相比并都不具有显著优势。

　　Cai et al. (2017）在Cai & Zhao (2016)的基础上，通过简化网络结构，混合字词输入以及使用早期更新（early update）等收敛性更好的训练策略，设计了一个基于贪心搜索(greedy search)的快速分词系统（见图4右）。该算法与之前的深度学习算法相比不仅在速度上有了巨大提升，分词精度也得到了进一步提高。实验结果还表明，词级信息比字级信息对于机器学习更有效，但是仅仅依赖词级信息不可避免会削弱深度学习模型在陌生环境下的泛化能力。表7列举了最近3年和速度相关的神经分词系统的结果。从中可见，Cai et al. (2017)首次使神经模型方法在性能与效率上同时取得了和传统方法相当的成绩。

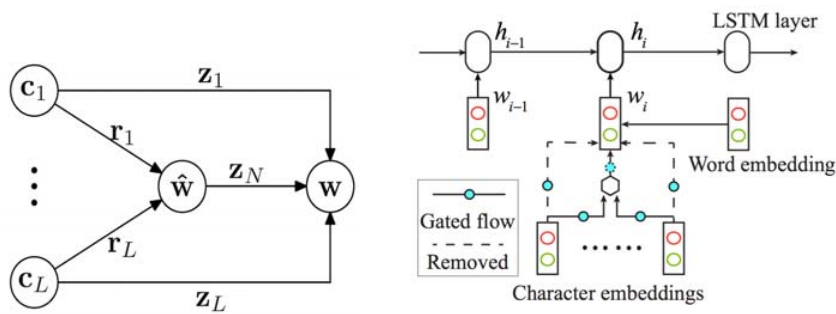

表4: Cai & Zhao (2016) (左)和Cai et al. (2017) (右)的组合门网络模块

## 4 封闭及开放测试

　　SIGHAN Bakeoff的分词评测定义了严格的封闭测试条件，要求不得使用训练集之外的语言资源，否则相应结果则算开放测试类别。区分封闭和开放测试的一个主要目的，是分辨机器学习的性能提升的确是模型自身的改进，而非其它。

　　不管是传统模型还是深度学习模型，可选的分词用外部资源都可以包括各类词典和切分语料（不一定和已有切分语料属于同一个分词规范）。外部资源的使用，可以通过额外标记特征的形式引入，早期的开放测试系统包括Low et al. (2005)。Zhao et al. (2010a)系统考察了多种外部资源，包括词典、命名实体识别器以及其他语料上训练的分词器，统一用于字标注模型下的附加标记特征，所提的具体做法很简单：在主切分器上加入其它分词器（或命名实体识别器）给出的辅助标记特征即可。结果表明，该策略在所有分词规范语料上都能显著提升性能，特别是在SIGHAN Bakeoff-2006的两个简体语料上可以带来额外的2个百分点的性能增益。表6展示的结果显示，Zhao et al. (2010a)报告的开放测试结果目前为止依然为业界最高的分词性能。该组结果实际上在Bakeoff-2006语料上给出，因而缺乏PKU上的结果，所用的附加资源则来自其它公开的Bakeoff语料。最后，该工作还经验性暗示，如果可用的额外切分语料可以无限制扩大，则分词精度也可以无限制提升，虽然代价是切分速度会急剧下降。

　　基于嵌入表示的深度学习模型对于分词的封闭和开放测试区分带来了新的挑战。显然，在外部预训练的字或者词嵌入向量属于明显的外部资源利用，因为字向量预训练可以直接借用外部无标记语料，典型如维基百科数据，而词向量的预训练则需要使用一个传统分词模型在外部语料上作预切分，这会同步地引入外部资源知识并隐性集成传统分词器的输出结果。但是，相当部分的神经分词的工作有意无意地忽略了以上做法的角色区分，实际上等于混淆了开放和封闭测试，更不用说很多神经模型系统甚至再次使用额外的词典标注来强化其性能。这些做法严重干扰了对于当前神经分词模型的分析和效果评估：到底这些模型声称的性能提升，是来自新引入的深度学习模型，还是属于悄悄引入的外部资源的贡献？从表6中所比对的神经分词器的开放和封闭测试效果可以看出，大部分神经分词系统引入外部辅助信息，才能再获得1-2个百分点的性能提升（已经属于开放测试范畴），才能和严格封闭测试意义上的传统模型抗衡。如果严格剥离掉所有额外预训练的字或词嵌入、额外引入的词典标注特征以及隐性集成的传统分词器的性能贡献，可以公正地看出，直至2016年底，所有神经分词系统单独运行时，在性能（更不用说在效率上）都不敌传统系统。

表6: SIGHAN Bakeoff-2005评估语料上的不同分词方法性能比较(F1值)

|  | 封闭测试 | | | | 开放测试 | | | |
| --- | --- | --- | --- | --- | --- | --- | --- | --- |
|  | PKU | MSR | CityU | AS | PKU | MSR | CityU | AS |
| Tseng et al.(2005) | 95.0 | 96.4 | 95.2 | 94.7 | | | | |
| Zhang & Clark (2007) | 94.5 | 97.2 | 94.6 | 94.7 | | | | |
| Zhao & Kit (2008b) | **95.4** | **97.6** | **96.1** | **96.5** | | | | |
| Sun et al. (2009) | 95.2 | 97.3 | 94.6 | 95.7 | | | | |
| Zhao et al.(2010a) | - | - | - | - | - | 98.3 | 97.8 | 96.1 |
| Sun et al.(2012) | **95.4** | 97.4 | 94.8 | - | | | | |
| Zhang et al.(2013) | - | - | - | - | | 96.1 | 97.4 | |
| Zheng et al.(2013)** | 92.4 | 92.8 | - | - | 93.3 | 93.9 | | |
| Pei et.al (2014) | 93.5 | 94.0 | - | - | 94.4 | 94.9 | | |
| Chen et.al (2015a) | 94.4* | 95.1* | - | - | (96.4) | (97.6) | | |
| Chen et al.(2015b) | 94.3* | 95.0* | - | - | **(96.5)** | (97.4) | | |
| Ma and Hinrichs (2015) | 95.1 | 96.6 | - | - | | | | |
| Cai and Zhao (2016) | 95.2 | 96.4 | - | - | 96.5 | 96.5 | | |
| Xu and Sun (2016) | | | - | - | (96.1) | (96.3) | | |
| Zhang et al. (2016) | 95.1† | 97.0† | - | - | 95.7 | **97.7** | | |
| Liu et al. (2016) | 93.9† | 95.2† | - | - | 95.7† | 97.6 | | |
| Yang et al.(2017) | | | | | 96.2 | 97.3 | **96.7** | **95.4** |
| Cai et al. (2017) | **95.4** | 97.0 | 95.4 | 95.2 | | | | |

说明：表格上部展示的是传统方法，下部是深度学习方法。标有双星号(**)的是来自Pei et al. (2014)的再运行结果；标有星号(*)的是来自Cai & Zhao (2016)的再运行结果；带有†是指使用了或者可能使用了预训练的字向量；带有‡侧是依赖传统模型(在大规模未标注语料上使用传统切分的结果进行预训练)；而括号(…)里的结果使用了成语表。

表7: 深度学习分词算法性能和效率的综合比较

| 模型 | PKU | | | | MSR | | | |
| --- | --- | --- | --- | --- | --- | --- | --- | --- |
| | $F_1$ +预训练 | $F_1$ | 训练（小时） | 训练（秒） | F1 +预训练 | F1 | 训练（小时） | 训练（秒） |
| Zhao and Kit (2008b) | - | 95.4 | - | - | - | 97.6 | - | - |
| Chen et al. (2015a) | 94.5* | 94.4* | 50 | 105 | 95.4* | 95.1* | 100 | 120 |
| Chen et al. (2015b) | 94.8* | 94.3* | 58 | 105 | 95.6* | 95.0* | 117 | 120 |
| Ma and Hinrichs (2015) | - | 95.1 | **1.5** | **24** | - | 96.6 | **3** | **28** |
| Zhang et al. (2016) | 95.1 | - | 6 | 110 | 97 | - | 13 | 125 |
| Liu et al (2016) | 93.91 | - | - | - | 95.21 | - | - | - |
| Cai and Zhao (2016) | 95.5 | 95.2 | 48 | 95 | 96.5 | 96.4 | 96 | 105 |
| Cai et al. (2017) | **95.8** | 95.4 | 3 | 25 | **97.1** | 97.0 | 6 | 30 |

说明标有星号(*)的数据来自Cai and Zhao (2016)再运行的结果。此表列出的是Zhang et al. (2016)与Liu et al. (2016)中神经网络模型单独工作的结果。注意大多数深度学习方法使用的字向量可以事先在大规模无标记的语料上进行预训练。严格来说，这类结果需归于SIGHAN Bakeoff开放测试的类别。

Yang et al. (2017)专门调查分析了外部资源对中文分词效果的影响，包括预训练的字/词向量、标点符号、自动分词结果、词性标注等，他们把每一种外部资源当作一个辅助的分类任务，使用多任务神经学习方法预训练了一组对汉字上下文建模的共享参数。大量的实验表明了外部资源对神经模型的性能的提升同样具有重要意义。

如果把外部资源的贡献进行量化，或者简化一些，是否能够给出机器学习的语料规模和学习性能的增长之间的联系规律？其实，这方面的经验工作已在Zhao et al. (2010b)之中完成，基本结论是统计机器学习系统给出的分词精度和训练语料规模大体符合Zipf律，即：语料规模指数增长，性能才能线性增长。而和统计分词不同，更传统的规则分词，例如最大匹配法，其精度和所用的词典（即所收录的词表词）的规模成线性关系，因为分词错误主要是未登录词导致的。这一结论意味着统计方法，无论是传统的字标注还是现代的神经模型，仍有着巨大的增长空间。

## 5 结论

关于中文分词的机器学习方法，长期以来一直存在着"字还是词"的特征表示优越性之争，这恰好和语言学界对于中文结构分析的"字本位"还是"词本位"的争议相映成趣。这一点早在黄和赵 (2006)中就给出了经验性观察结果：字、词的特征学习需要在分词系统中均衡表达，才能获得最佳性能。实际上，所谓字、词争议的核心对应于分词的两个指标，已知词（或词表词，即出现在切分训练语料中的词）的识别精度和未登录词的识别精度，前者识别精度很高、相对容易但所占百分比高，后者识别精度很低、难度较大但所占百分比比较低。经验性的结果表明，强调基于字的特征及其表示会带来更好的未登录词的识别性能。原因无他，未登录词从未在训练集出现，只能依赖于模型通过字的创造性组合才能识别。反过来，强调词特征的系统，包括基于词的切分系统，对于未登录词的识别效果通常略为逊色。最佳的分

词系统总是需要合理考虑字表示和词表示的平衡问题。最近的两个工作的改进点可以辅证这一结论：Cai et al. (2017)对于Cai & Zhao (2016)的一个关键性改进，是词向量不再总是由字向量通过神经网络计算得到，而是采取了两种策略，即低频或者未知词继续由字向量计算，而训练集中的高频词（可以认为是更为稳定的已知词）则进行直接计算。当系统由后者偏向字向量表示的模式转向字-词均衡的表示模式以后，确实带

来了额外的性能提升。

最近5年，基于神经网络模型的分词学习已经取得了一系列成果。就目前的结果来看，我们可以得出两个基本结论：一，神经分词所取得的性能效果仅与传统分词系统大体相当，如果不是稍逊一筹的话；二，相当一部分的神经分词系统所报告的性能改进（我们谨慎推测）来自于经由字或词嵌入表示所额外引入的外部语言资源信息，而非模型本身或字词嵌入表示方式所导致的性能改进。如果说词嵌入表示蕴含着深层句法和语义信息的话，那么，这个结论似乎暗示一个推论，即分词学习是一个不需要太多句法和语义信息即可良好完成的任务。

现代深度学习意义下的神经网络归类于人工智能的联结主义思潮，由于其带有先天性的内在拓扑结构，如果能克服其训练计算低效的弊病，它就应该是本身需要结构化学习的自然语言处理任务的理想建模方式。这是我们在深度学习时代看到更多样化的结构建模方法用于中文分词任务的主要原因。如果我们能有效平衡字-词表示的均衡性，不排除将来深度学习基础上的分词系统能有进一步的成长空间。

## 参考文献

**[作者简介]**
赵海，男，上海交通大学计算机科学与工程系博士、副教授，从事计算语言学等教研，多次获自然语言处理（包括中文分词等）国际评测第一名。
蔡登，男，上海交通大学计算机科学与工程系硕士研究生。
黄昌宁，男，教授、高级研究员，国内计算语言学奠基者之一，是清华大学计算机科学与技术系和亚洲微软研究院两处学术重镇的自然语言处理组创始人，有众多学生为学界翘楚。
揭春雨，男，副教授，清华大学计算机科学与技术系毕业，谢菲尔德大学计算机科学博士，目前香港城市大学翻译及语言学任教，博、硕士导师，获终身教职，主要从事计算语言学和术语学等教研工作。